\documentclass[oribibl]{llncs}

\usepackage{graphicx}
\usepackage{epsfig}
\usepackage{url}

\title{Towards Anthropo-inspired Computational Systems: the $P^3$ Model}

\author{
Michael W. Bridges \inst{1}
\and
Salvatore Distefano \inst{2, 6}
\and
Manuel Mazzara \inst{3}
\and
Marat Minlebaev \inst{4,5,6}
\and
Max Talanov \inst{6}
\and
Jordi Vallverd{\'u} \inst{7}
}

\institute{
iCarnegie Global Learning, USA.
\and
Politecnico di Milano, Italy.
\and
Innopolis University, Russia.
\and 
Institut de Neurobiologie de la M{\'e}diterran{\'e}e, France.
\and
Aix-Marseille University, France.
\and
Kazan Federal University, Russia.
\and
Universitat Aut{\`o}noma de Barcelona, Catalonia, Spain.
}

\titlerunning{Towards Anthropo-inspired Computational Systems}
\authorrunning{M. Bridges, S. Distefano, M. Mazzara, M. Minlebaev, M. Talanov \& J. Vallverdu}

\begin{document}
\maketitle
\begin{abstract}
This paper proposes a model which aim is providing a more coherent framework for agents
design. We identify three closely related anthropo-centered domains working on separate functional
levels. Abstracting from human physiology, psychology, and philosophy we create the $P^3$  model to be
used as  a multi-tier approach to deal with complex class of problems. The three layers identified in this model
have been named PhysioComputing, MindComputing, and MetaComputing. Several  instantiations of this model
are finally presented related to different IT areas such as artificial intelligence, distributed computing, software and
service engineering.
\end{abstract}

\section{Introduction}\label{introduction}
 
Emulation of functional aspects of living systems have been so far based on the perspective of
linear interaction between sets of pre-programmed subsystems. As a consequence, the exhibited behavior
is that of modular systems respectively responsible for specific functional domains.
 
%Currently all computational attempts to imitate or emulate functional aspects of living systems
%have been based on a perspective which considers a linear interaction between sets of programmed code.
%Thus, those systems behave (sometimes) as multilayered systems that work with modules reigning over
%specific functional task domains.
 
Development of cognitive science has recently shown as living beings actually integrate several functionalities
which operate in parallel and emergence of higher functions like consciousness may not be possible without the
complex cooperation of such sub-systems.
 
%From a grounded cognition approach that includes recent views on cognitive systems as embodied,
%situated or enactive entities, a living being integrates several parallel functionalities
%that one running at more complex levels make possible the emergence of things like consciousness.
 
In this paper, we introduce a three domain model providing a computational framework for an ecosystem 
of cognitive entities. This approach is then extended to computing systems and then instantiated
to specific IT domains such as robotics, services or networks. The model has to be seen as a multi-views
representation of phenomena of like artificial consciousness, artificial creativity and artificial intuition.
Each view is best illustrated by an analogy with well-known anthropocentric domains: physiology, psychology, philosophy ($P^3$).
 
%The $P^3$ model is a novel approach inspired by combination and interaction of the philosophy,psychology and physiology of the anthropocentric perspective of human body/brain/mind and metaconcepts.
 
The framework can be seen as a set of perspectives in order for researchers to grasp a better
understanding of a specific phenomena, starting from the lower level (i.e. hardware and software implementation)
up to the higher (i.e. artificial intelligence (AI) or cognitive robotics).

\section{The $P^3$ anthropocentric model}\label{the-anthropo-centric-model-underlying-the-new-domains}

Human brain  follows a non centralized subsumption architecture in which consciousness is a very important but not determinant managing layer. We must operate into dynamical environments out of our full control. It explains why, according to their internal states, memories or  necessities, people faced to similar environmental conditions react differently. It can be justified by the fact that the topology of cognition is distributed all along the body, the cultural sphere and the brain. Up today AI attempts to reproduce or emulate human intelligence have failed in the achievement of similar adaptative and multilayered systems. Our aim is to produce more efficient, problem-solving and innovative AI systems providing a new architecture design.

\begin{figure}[htbp]
\centering
\includegraphics[width=0.5\textwidth]{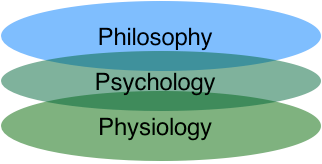}
\caption{The $P^3$ model. \label{fig:hlmod}}
\end{figure}

We consider three domains as the real modules of any truly intelligent system which produces complex behaviors, reasonings and belong a consiousness, as shown in Fig. \ref{fig:hlmod}, bottom up: physiology, psychology and philosophy of $P^3$ model.
This is the starting point towards a new approach where we consider these domains as layers of a layered model. Each layer implements specific functionalities and interacts with the adjacent ones through specific interfaces. 
 
\subsection{The Layers}

\subsubsection{Physiology}
\label{sec:physio-layer}

Physiology is the study of basic normal function in a living system, focusing on organisms
and on their main physical functions \cite{HALL}. The subdomain of physiology that studies
the central nervous system is neuroscience. The principal functional unit of our brain is neuronal cell, i.e. the ``brick'' of central nervous system. %\cite{neuron}. 
The principal difference of the neuron from major part of the cells is ``excitability''.
That as well as an option to transfer the information between the neuronal cells allow to transmit the signal along multiple cells. The intercellular connections are selforganised that and neuronal chains form networks that serve specific particular functions. The simplest example is knee reflex, when light kick of the knee excites the receptor part of the sensory neuron that carries the sensory signal to the spinal cord, where it evokes the activity in intermediate cell that in its turn activates the motor neuron making the contraction of the hip muscles. 

Thus, the neuronal network is an organized structure of the neurones, that serves specific function, i.e. ``brick wall''. %\cite{Biological_neural_network}. 
However, presence of multiple factors i.e. different cell classes (inhibitory and excitatory) and diversity of the neuronal interconnections (axosomatic, axoaxonal, perisomatic, axodendritic etc) and other factors (such as extracellular medium content, that affects electrical properties of the cells, neuronal plasticity
of the synaptic strengths based on the previous experience of the cell) provides the enormous freedom for the functionality of our neuronal system for decision making even at the microscopic level. Considering the entire brain, the direct measurements of the immediate state of the small neuronal networks allow us to describe the general principles of the organisation of the central nervous system and its functioning but are hardly sufficient for analysis and prediction of the results of the concrete complex task, like decision making. We have to take into account the interactions between multiple functionally linked structures of the brain, previous experience (including acquired skills) and the temporary the state of mind
to do that. This in its turn places us into completely another domain of science investigating our thinking brain
--- psychology.

\subsubsection{Psychology}\label{psychology-layer}

Psychology studies the behavior and mental activity of human beings  according to the brain processes with which are correlated \cite{Matsumoto} . Psychologists cover a broad range of interests regarding mental processes or behaviors: from the human being as an individual being(internal data), as well as considering the human as a node of a web of social, collective and ecological interactions
(external). Although cultural variables are very important in order to understand the role of psychic events into human actions \cite{Nisbet} , we can affirm that psychology deals more directly with the several brain data processing mechanisms that operate at a presymbolic level. Consequently aspects like attention, visual field selection,
interaction management, time perception, mental disorders, among others, belong to the field of this research area. Other very important  mechanisms that fall under the general category of mental processes are attentional mechanisms, decision-making mechanisms, language processing mechanisms or social processing mechanisms\cite{Putnam}. 
%From wikipedia 
%We could add that 
As stated above, Psychology is not directly involved into the neural mechanisms that are present during mental actions,but on the functional mental managing of processed data. Let us use a metaphor: if we take brain as hardware, then psychology is the operating system, and philosophy is the software. Psychologists work with the ways brain transforms raw data into informational data, from several domains: cognition (thought), affect (emotion),
and behaviour (actions).  In combination, these fundamental units create human experience.

For instance, the visual sensory system may detect a large, dark, fury shape approaching quickly. Through a broad range of psychological processes and mechanisms, the shape might be interpreted as a viscous dog about to attack.  The output would be a mental representation of a dangerous animal (cognitive), as well as trigger an emotion like fear (affect) and, finally, elicit a response such as running behavior).  Alternatively, based on prior experience with this animal, the output might be a mental representation
of a friendly animal (cognitive), which provokes an excitement (affect) and ask the human for a specific action like petting (behavior).  The sensory input is the same, but the processing results in very different outputs.
Highest level perspective of the psychology domain are formed of the concepts describing whole mind states and transitions like: ego and super-ego, personality types, tempers and so on that brings us closer to psychologically philosophical question ``Who am I'' and philosophy of cognition.

\subsubsection{Philosophy}\label{philosophy-layer}

Philosophy is the study of general and fundamental problems, such as those connected with reality, existence, knowledge, values, reason, mind, and language \cite{Philo}. Philosophy is distinguished from other ways of addressing such problems by its critical, generally systematic approach and its reliance on rational argument, without supranaturalist influences  \cite{epistemology}. Despite of specific cultural differences between Eastern
and Western philosophical paradigms (whose boundaries are not so clear across regions and periods), philosophy looks for the rational analysis of reality and the obtention of metainformation.
Philosophers not only know that they know, but also ask themselves how is it possible,
which are the consequences of this process and other related questions. During this process,
philosophers create heuristics of knowledge adapted to the surrounding socio-epistemic circumstances. This implies the work at the meta-level analysis. Our use of the concepts like: consciousness, emotions, mind, intuition, talking about a specific computational system domain, deals with the necessity of a meta-level of analysis. This implies to own the capacity of creating cognitive tool; it allows to extend the processing of information from the brain (and body) to external devices. Thus, humans are extended entities. And they evolve, culturally, creating meaning and ways to obtain meaning. According to the previous explained notions, philosophy is a meta-level of data acquisition and processing. It's sphere belongs to the work at symbolic level, using semantic
and syntactic mechanisms in order to create new information.
 
 \subsection{Integrating domains}\label{integration-layer}

Previous sections sketched our holistic way to understand the multilayer nature of a cognitive process. 
The living and cognitive entity is based onto a physiological structure (the body) that
by virtue of its specific architecture determines a strict way to interact with the environment
as well as forces to follow the requirements of its own bodily system. Inside this body, several mechanisms
deal with internal and external sets of data that must be analyzed in real time to produce adequate responses,
if the organism that wants to survive and fulfil its intentional necessities (energy conservation,
energy acquisition, data communication --- as reproduction or cultural transmission ---,
maintenance of the system --- playing, exercising, etc). 

This internal process of data management is handled by the brain, at the psychology level,
and fully modelled and controlled by emotional mechanisms that allow the evaluation of acquired data as well as it's storage and application during decision processes. This allows to this entity to create a semantic view of the world and a second-order intentional approach to reality. At this point, all the processes are the result of a self-emergent data integration mechanism which makes possible the emergence of the binding activity called ''consciousness''. But it is still a passive process because it lacks a symbolic way to understand it, which leads us to the next level: philosophical.  

Finally, this system creates symbolic ways to perform natural calculations. These bodily calculations are initially inspired by bodily requests such as calculating amounts or sets of objects (subitization, numerosity)or predicting possible outcomes of actions and events (the cause of events, the consequences of special events such as death, survival constraints, malfunctioning of the system,...). From this binding process at several layers energes a metalevel: the conscious Self. It is a mechanism to integrate more efficiently and dynamically strategies of actions based on bodily and psychological constraints but also connected socially with other entities. At this level we enter into the realm of meta-information, created by the entity using externally shared tools like symbols, concepts or reasoning strategies/heuristics. 

This architecture explains how the  layers identified in the $P^3$ model interact to produce an activity,which is always implemented as a combination of actions, operations and interactions at the three layers and among them. 
%related. 
The sensorimotor leads to psychological processing, which affects the conscious mind, although from a functional perspective the three domains are independent and directly fully controlled by none of the domains. They affect themselves but each one reigns on a different informational level.

\subsection{Abstraction}\label{abstraction-1}
The $P^3$ model can be abstracted from a human perspective into a more general system view.
This way, the physiology layer, at the bottom, is  mainly focused on the \emph{main (physical) mechanisms}
implementing and providing the \emph{basic (mandatory) functionalities} required for the system operation.
These basic functionalities are used at higher level to implement more complex functions, policies and strategies.
Specifically, the psychology layer mainly implements individual reasoning and behaviour by just taking into account
an \emph{introspective} (internal, from the inside) view.
In the system view this corresponds to advanced functionalities and policies operating \emph{locally}
to a system node or component, taking into account this local perspective.
On the other hand, philosophy provides a wider, \emph{introspective} or external view, by which the individual
reasons from an external perspective as just an element of a wider global context, a kind of estrangement
characterizing a meta-layer/view.
This way, from the system perspective, the philosophy layer provides advanced functionalities and policies
operating at a \emph{global} view, considering the overall systems and its interaction with the external environment.

Highlighting both horizontal and vertical relationships different interpretations are possible:
any block (physiology, psychology, philosophy) is independent, and therefore independently mapped
into the application domain, or as a whole. Layers are connected and interdependent in layered way,
complex functions at higher layer are based on simpler ones provided at lower layer(s).

The following example describes this layered model and its abstraction in the neuroscience area.

\subsubsection{Physiology}\label{neuroscience}
Example of high level abstraction of this domain could be artificial living
organism or artificial living spiking pseudoneuronal network. This implies artificial life as main
concept of each building block of the system. On the other hand main
property of artificial living system could be the adaptation,
self-evolving, and self-organizing abilities that defines overall
behaviour of the system. The lowest level abstraction is
artificial living cell or an object. Artificial living cell (ALC) ---
basic element of the system with self adaptation mechanisms to ever
changing environment in some boundaries. One of the types of ALC is
artificial pseudoneuron --- mainly targeted on management and information processing of the
artificial living organism (ALO). ALC should be self adaptable for the
environmental changes in some boundaries. Mid level is represented by
artificial organs or devices that are created via ALCs in form of ALC
networks. An artificial living organism (ALO) is the combination of
artificial organs which provide the functions set necessary for the
artificial life.

\subsubsection{Psychology}\label{psychology}

Most of complex computational heuristics run by AI experts deal at a certain level with the implementation
of processes that could be labelled as 'psychological' if they were run by human brains. 
We are thinking of heuristics of decision under poor informational environments, learning processes
(supervised or unsupervised), artificial creativity, artificial vision, semantic data mining, etc.
But in no case, these implementaions have been done once clarified the main architecture of the system,
as we've been doing now. Cognitive architectures like SOAR, CHREST, CLARION, ICARUS, DUAL, and Psi are based
on a full symbolic level \cite{computationalmodelsemotion, computationalmodelsemotionscognition}. 

There is not a bottom up increasing complexity that justifies the interaction between the hard-physio
and the upper levels of data processing (like psychological or also meta-heuristic ones). Nevertheless,
the cited cognitive architectures have succeeded on simulating specific functionalities of the human mind.
A good approach should be grounded as well as embodied, allowing a true merging of layers of interaction
of a system with their environment as well as with other intelligent systems.

\subsubsection{Philosophy}\label{philosophy}

\emph{Philosophical corollaries about thinking}

Philosophical research has not an unit of study. Historically, at least
for Western thought, the syntax coherence of written thoughts was
considered the basis of philosophical analysis; that is, logical
coherence, from Aristotle's syllogistic to contemporary non-monotonic or
fuzzy logics. But also in parallel, there is another stream of studies
devoted to the internal feeling about the world (phenomenology) and the
corresponding debates on the existence or not of a perceiving unit
(\emph{I, Me, Myself}). Curiously, with the recent possibility of
analyzing neural correlates \emph{in vivo}, the internal states of mind
are now partially discretizable. In both cases, the epistemic and
experiential relationship with reality belong to a symbolic level of
action. And in all philosophical systems is studied where are the limits
of the own system and how we can know that our knowledge is correct. In
a nutshell: it implies a meta-analysis level. Is in that sense that the
philosophical level can be understood from two different perspectives:
as a specific way to think on the value of symbolic processes as well as
ac activity to think about the own existential experience (``to be
conscious of''). Our aim is not to create artificial consciousness or
artificial systems that think about their existence at this point but to
create possible mechanisms of innovative ways to deal symbolically with
information. This is the meta-level value of the notion of philosophy
with which we are working to.

This meta-skill is mainly defined in the philosophy layer and 
%what we have defined the MetaComputing domain. 
we think on how meta-levels of significance are achieved and the mechanisms by
which humans improvise new ways to understand and define their relevant
aspects of reality.

\section{A $P^3$  Computing System Model}\label{toward-a-mapping-to-computer-systems}

\begin{figure}[htbp]
\centering
\includegraphics[width=1.0\textwidth]{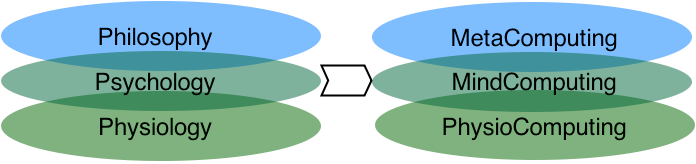}
\caption{The $P^3$ computing model.\label{fig:compmap}}
\end{figure}

Starting from the $P^3$ anthropocentric model of Fig. \ref{fig:hlmod} in this section we map the three layers there identified into the corresponding layer for an abstract computing system. 
From our perspective, a computing system is a physical and/or abstract system, such as computer machines, devices, robots, networks, distributed systems, software, services, algorithms, workflows, data, in computer science and engineering.
This way, a new, paradigmatic approach for computing is specified in Fig. \ref{fig:compmap} where three  layers
%new scientific domains 
are identified:
MetaComputing, MindComputing, PhysioComputing. 
%that should incorporate different views on severalcognition related phenomena: emotions and motivation, common sense logic, memory, consciousness, awareness, learning, anticipation, subjective experience, intuition, perception and understanding, creativity, dream/sleep.
These layers  interact  each other and implement  functionalities of the corresponding $P^3$ model layers, i.e. physiology, psychology and philosophy, respectively, keeping relationships, interactions and basic mechanisms of the former model, just specialized in computer systems
%explain phenomena listed above from different
%perspectives and layers of abstraction and granularity.
%In our model PhysioComputing, MindComputing and MetaComputing are deeply interconnected. 

\subsection{PhysioComputing}\label{physiocomputing}

PhysioComputing implements the \emph{physiology of computing}. 
It is therefore usually related to
%the scientific study of 
the hardware and the low level software, e.g. the firmware, providing the basic functionalities for a computing system to properly work.
%intelligent systems/approaches
%that are capable of the set of phenomena 
%listed above 
%of  artificial living systems (ALS). 
%It should take into account
%implementation details like distributed computing systems or supercomputer systems and 
%it should take into account
%cellular analogies with human organism of vital functions, life and existence in the everchanging realistic environment.
From an abstract point of view PhysioComputing provides the basic mechanisms implementing the main functionalities for a computing system to work.
The scope of PhysioComputing is on structural, functional, organizational and
%device-embedded, distributed, 
communicational aspects
of computing systems. 
%in a device, chip, service, software etc with a software that creates proper collaboration intellectual system operations.

\subsection{MindComputing}\label{mindcomputing}

MindComputing is the \emph{psychology of computing}. 
%is an academic study of a phenomena of ALS that implements main components
%of the artificial intelligent systems analogous to natural intelligent systems like mind. 
It should consider the high level details of phenomena taking into account broader and more conceptual approaches
of the artificial mind operation based on the concepts, mechanisms, and functionalities
%theories/hypotheses 
introduced and provided at lower level by the PhysioComputing.
From an  abstract computer systems perspectives, PhysioComputing implements advanced and enhanced, mechanisms, policies, and strategies for locally optimizing the system, just considering the system introspectively.
Scope of the MindComputing is therefore to implement enhanced services or systems
% cognitive functions of ALS, social collaborations and behaviour of artificial individuals exploring 
on top of basic mechanisms, processes and functionalities provided by the lower level.
%that underlie cognitive functions. 

\subsection{MetaComputing}\label{metacomputing}

MetaComputing is the \emph{philosophy of computing}.
%study of fundamental problems of an ALS. 
%It should use several views possibly 
Based on the MindComputing and PhysioComputing concepts, mechanisms and policies, it provides a further abstraction from a more general viewpoint, i.e. considering the interaction of the overall computing system with the environment, from outside.
%which could be integral part of the Philosophy.
%The creative ways obtaining new knowledge (this includes information management, strategies, ways to obtain raw data) are considered on this level. 
%As humans work inductively, abductively, statistically, improvising in different moments and situations, this implies the necessity of the exploration of cognitive-knowledge meta-levels.
This is a meta-level, where the computing system is considered as part of the environment and its interactions with the other systems or elements are taken into account.
The scope of MetaComputing includes definitions and high level, meta- views on the problems, thus providing meta-solutions to be enforced as policies and mechanisms at lower  MindComputing and PhysioComputing layers.
%of ALS (organisms), artificial individuals/personalities as part and extending the Artificial Intelligent domains.
%In other words, MetaComputing represents a methodological challenge that could be solved
%thanks to emotional cognitive architectures.

\section{Applications}\label{applications}

To get the narration to more practical perspective we provide several examples of possible applications of introduced above framework $P^3$. We describe social network and cognitive robotics domains represented in the three layered perspectives of philosophy, psychology, physiology. This trifocal view on the complex phenomena could be beneficial in several domains that are demonstrated below.

\subsection{Social Network}\label{sec:sn}

One of the most significant applications of the concepts described so far is in the social networks domain. Internet and the Web constitute a global artificial organisms/ecosystem on which actual life, in the broad sense we intend it, can be built. At the \textit{physiology (PhysioComputing)} level this organism (node of the network) represents the basic ``building block'' of the living ecosystem that can be built on top of it. Applications like Facebook allow users to create profiles and self-identification in the living organisms. Facebook profiles precisely represent, at the \textit{psychology (MindComputing)} level, the subjective self characterization of users in relation to other individuals populating the network. The metaphor solidly applies to the \textit{philosophy (MetaComputing)} level too, where meta-analysis on the organism activities is performed. Here is where, for example, analysis of big data or opinion mining stands. The system has the peculiar ability to reason about itself at this level, and this can be done by the users themselves\cite{Mazzara2013}. 

\subsection{Cognitive robotics}\label{robotics}

One more application of MetaComputing, MindComputing and PhysioComputing to the cognitive robotics domain invokes three new emerging domains: MetaBotics, MindBotics, HardBotics. Where in HardBotics we could identify following concpts: artificial living systems that cold be capable of reproduction thus regeneration via universal living ``bricks/cells''. Specifically neuronal systems could be capable of reconfiguration of their connections/``synapses'' and generation of new neurons via ``artificial neurogenesis''. MindBotics contains definitions of several phenomena: affects, high-level emotions,temper, psychotypes, consciousness etc. In its turn a robotics system could fit in to the philosophical ``model of 6''\cite{emotionmachine} by Marvin Minsky using emerging effects of living ``artificial bricks/cells'' like pattern matching, predictions, associative learning, deliberations, reflections, self-consciousness. Thus trifocal approach could be mapped and used for the benefit of robotic systems providing extended reflections of phenomena and their processes for more complete and exhaustive picture of robotics systems and their environment. 

\section{Conclusions and Future Works}\label{conclusions-and-future-works}
 
The basis of any biological system is to be alive and keep living. Viruses (for which
is still under debate whether they belongs to the realm of living entities), also have
feeding and reproduction as their fundamental aims. Once these systems acquire tools to
process bigger arrays of data, they can create more complex patterns of interaction.
This base is well understood by neurobiologists, but is hard to implemented into current
AI applications.
 
The approach presented in this work aims at providing a more coherent framework for agents
design identifying three closely related domains, though working at separate functional levels.
 
\begin{itemize}
                \item The foundation of artificial living system stands in the idea of \emph{PhysioComputing};
                \item Self-emergent characters, based on self-organization of connections and signals as processed by living cells
                (neurons) is the main aim of \emph{MindComputing}. Consciousness, for example interpreted  as a self-emergent property
                of neuronal networks;
                \item From a functional perspective, consciousness is the result of a multi-integration of data, but the feeling of being
                performing a conscious experience relies completely on the symbolic level providing tools that create elaborated semantic
                frameworks. Symbolic processing lies a basement for \emph{MetaComputing}.
\end{itemize}

\bibliographystyle{splncs03}
\bibliography{three_domains_intro}

\begin{thebibliography}{10}
\providecommand{\url}[1]{\texttt{#1}}
\providecommand{\urlprefix}{URL }

\bibitem{HALL}
Hall, J.: Guyton and Hall textbook of medical physiology (12th ed. ed.).
  Saunders/Elsevier (2011)

\bibitem{computationalmodelsemotionscognition}
Lin, J., Spraragen, M., Zyda, M.: Computational models of emotion and
  cognition. Advances in Cognitive Systems  2,  59--76 (2012)

\bibitem{computationalmodelsemotion}
Marsella, S., Gratch, J., Petta, P.: Computational models of emotion. In:
  Scherer, K., Bänziger, T., Roesch, E. (eds.) A blueprint for a affective
  computing: A sourcebook and manual. Oxford: Oxford University Press (2010)

\bibitem{Matsumoto}
Matsumoto: The Cambridge Dictionary of Psychology. Cambridge University Press
  (2009)

\bibitem{Mazzara2013}
Mazzara, M., Biselli, L., Greco, P.P., Dragoni, N., Marraffa, A., Qamar, N.,
  de~Nicola, S.: Social networks and collective intelligence: A return to the
  agora. In: Social Network Engineering for Secure Web Data and Services. IGI
  Global (2013)

\bibitem{emotionmachine}
Minsky, M.: The Emotion Machine: Commonsense Thinking, Artificial Intelligence,
  and the Future of the Human Mind. Simon \& Schuster (2007)

\bibitem{Nisbet}
Nisbet, R.: The Geography of Thought. NY: The Free Press. NY: The Free Press.
  (2003)

\bibitem{Putnam}
Putnam, H.: Reductionism and the nature of psychology. In: Haugeland, J. (ed.)
  Mind Design: Philosophy, Psychology, Artificial Intelligence, pp. 205--219.
  MIT Press, Cambridge, MA (1981)

\bibitem{Philo}
Rationality: The Cambridge Dictionary of Philosophy. Cambridge University Press
  (1999)

\bibitem{epistemology}
Tomberlin, J. (ed.): Philosophical Perspectives 13: Epistemology. Blackwell,
  Oxford (1999)

\end{thebibliography}
\end{document}